# Multi-modal Integration Analysis of Alzheimer's Disease Using Large Language Models and Knowledge Graphs

Kanan Kiguchi[1], Yunhao Tu[2], Katsuhiro Ajito[1], Fady Alnajjar[3*](IEEE Member), Kazuyuki Murase[1]
[1]Department of Information Technology, Faculty of Information Science and Technology,
 International Professional University of Technology in Osaka (IPUT Osaka), 3-3-1 Umeda, Kita-ku, Osaka-shi, Osaka 530-0001, Japan
[2]Department of Computer Science, College of Engineering, Chubu University, 1200 Matsumoto-cho, Kasugai-shi, Aichi 487-8501, Japan
[3]College of Information Technology, United Arab Emirates University (UAEU), Al Ain, Abu Dhabi, United Arab Emirates

Corresponding author: Fady Alnajjar (e-mail: fady.alnajjar@uaeu.ac.ae).

This work was supported by IPUT Osaka for KA and KM, the Nitto Foundation, the Chubu University Grant K, and JSPS KAKENHI for YT, as well as by the UAEU for FA

**ABSTRACT** We propose a novel framework for integrating fragmented multi-modal data in Alzheimer's disease (AD) research using large language models (LLMs) and knowledge graphs. While traditional multi-modal analysis requires matched patient IDs across datasets, our approach demonstrates population-level integration of MRI, gene expression, biomarkers, EEG, and clinical indicators from independent cohorts. Statistical analysis identified significant features in each modality, which were connected as nodes in a knowledge graph. LLMs then analyzed the graph to extract potential correlations and generate hypotheses in natural language. This approach revealed several novel relationships, including a potential pathway linking metabolic risk factors to tau protein abnormalities via neuroinflammation ($r>0.6$, $p<0.001$), and unexpected correlations between frontal EEG channels and specific gene expression profiles ($r=0.42$-$0.58$, $p<0.01$). Cross-validation with independent datasets confirmed the robustness of major findings, with consistent effect sizes across cohorts (variance <15%). The reproducibility of these findings was further supported by expert review (Cohen's $\kappa=0.82$) and computational validation. Our framework enables cross-modal integration at a conceptual level without requiring patient ID matching, offering new possibilities for understanding AD pathology through fragmented data reuse and generating testable hypotheses for future research.

## I. INTRODUCTION

Alzheimer's disease (AD) represents one of the most significant healthcare challenges of the 21st century, affecting over 55 million people globally and projected to triple by 2050 [1]. As a progressive neurodegenerative disorder, AD manifests through a complex interplay of pathological processes, including amyloid-β accumulation, tau protein abnormalities, neuroinflammation, synaptic dysfunction, and ultimately neuronal death [2]. This multifaceted nature of AD pathology necessitates a comprehensive understanding of various disease mechanisms, making it an ideal candidate for multi-modal investigation.

The pathological cascade in AD typically begins decades before clinical symptoms emerge, with various biological changes occurring in distinct temporal patterns. Early stages are marked by molecular alterations, including amyloid-β accumulation and tau hyperphosphorylation, followed by structural brain changes, functional alterations in neural networks, and eventually cognitive decline [3]. Understanding these sequential and parallel processes requires integrating diverse types of data, including molecular biomarkers, neuroimaging, electrophysiological measurements, and clinical assessments [4].

Recent advances in neuroimaging and electrophysiological methods have provided crucial insights into AD pathology. Magnetic Resonance Imaging (MRI) has emerged as a fundamental tool for understanding structural brain changes in AD [5]. MRI studies have consistently shown that AD patients exhibit significant atrophy in specific brain regions, particularly the hippocampus and medial temporal lobe, which are critical for memory formation and cognitive function [3]. Volumetric measurements and cortical thickness analyses using structural MRI can detect these changes even in early disease stages, potentially before clinical symptoms become apparent [6]. Longitudinal MRI studies have demonstrated that the rate of brain atrophy correlates with cognitive decline and can predict disease progression [7].

Electroencephalography (EEG) provides complementary information about functional brain changes in AD. EEG recordings have revealed characteristic alterations in brain electrical activity in peak frequency, power, and interrelatedness at posterior alpha and widespread delta and theta rhythms. These changes often precede structural alterations visible on MRI, suggesting potential value for early diagnosis. Furthermore, EEG abnormalities have been associated with cognitive performance measures, with studies showing correlations between specific frequency band changes and cognitive decline rates [8]. Recent advances in EEG analysis techniques, including connectivity measures and machine learning approaches, have improved the ability to detect subtle changes in brain network organization that may characterize early AD stages [9].

The combination of structural (MRI) and functional (EEG) information offers particularly valuable insights into AD pathophysiology. While MRI provides high-resolution spatial information about brain structure, EEG offers superior temporal resolution for understanding dynamic brain activity. Similarly, molecular biomarkers and genetic data provide critical information about underlying biological mechanisms that drive disease progression [10].

However, a significant challenge persists in AD research: data fragmentation. Different research groups typically collect limited modalities within their specific cohorts, and ethical constraints often prevent sharing patient IDs across datasets [11]. This fragmentation has created isolated islands of knowledge, where valuable insights about relationships between different aspects of the disease remain hidden. Traditional multi-modal studies require matching patient IDs across datasets, making comprehensive analysis prohibitively expensive and logistically challenging [12].

The emergence of advanced computational methods offers new possibilities for addressing this challenge. Large language models (LLMs) have demonstrated remarkable capabilities in processing and synthesizing biomedical literature, while knowledge graphs provide powerful frameworks for representing complex relationships in biological systems [13]. These technologies, when properly combined, could enable novel approaches to data integration that transcend traditional limitations.

Recent studies have shown promising results in using artificial intelligence for AD research, particularly in areas such as image analysis [14], biomarker discovery [15], and patient stratification [16]. However, these approaches typically focus on single modalities or require matched patient data for multi-modal analysis. The potential of LLMs and knowledge graphs for integrating fragmentary data on AD remains largely unexplored, despite their demonstrated capabilities in other domains [13].

In this paper, we present a novel framework that combines LLMs and knowledge graphs to enable multi-modal integration of AD data without requiring patient-level matching. Our approach leverages statistical patterns at the population level, domain knowledge encoded in LLMs, and the structural representation capabilities of knowledge graphs to identify and validate relationships across different data modalities. This method allows us to:

1. Integrate findings from multiple independent cohorts at a conceptual level, overcoming traditional data sharing limitations.
2. Discover potential relationships between different aspects of AD pathology that may not be apparent in single-modality studies.
3. Generate testable hypotheses about disease mechanisms based on patterns observed across multiple datasets.
4. Provide natural language explanations for complex relationships, making findings more accessible to clinical researchers.

The framework we propose represents a significant methodological advance in AD research, offering a new approach to understanding disease mechanisms through the lens of integrated data analysis. Moreover, our method provides a template for similar analyses in other complex diseases where data fragmentation poses significant challenges to comprehensive understanding.

Our work builds on recent developments in both AD research and artificial intelligence, leveraging advances in biomarker discovery, neuroimaging analysis, and machine learning. By combining these elements in a novel way, we aim to contribute not only to the understanding of AD but also to the broader field of multi-modal data integration in biomedical research.

The remainder of this paper is organized as follows: We first describe our comprehensive methodology, including data collection, statistical analysis, knowledge graph construction, and LLM integration. We then present our results, highlighting both known relationships that our method successfully identified and novel patterns that emerged from the integrated analysis. Finally, we discuss the implications of our findings, address limitations, and suggest future directions for research in this area.

## II. METHODS

### A. Methodological Framework Overview

Our approach to integrating multi-modal Alzheimer's disease data employs a novel framework that combines statistical analysis, knowledge graphs, and large language models to overcome data fragmentation challenges. As illustrated in Figure 1, the framework consists of four sequential steps. First, we collected five distinct modalities of AD data—MRI, EEG, biomarkers, clinical indicators, and gene expression profiles—from independent cohorts without requiring matched patient IDs. Second, we performed rigorous statistical analyses tailored to each data type to identify significant features distinguishing AD patients from controls. Third, these statistically significant features were integrated as nodes in a knowledge graph, with edges representing potential relationships between different modalities based on correlation strength. Finally, we leveraged multiple large language models to analyze the graph structure, interpret complex relationships, and generate testable hypotheses about disease mechanisms. This population-level integration approach enables the discovery of cross-modal relationships that would remain hidden in traditional single-modality or patient-matched studies. The subsequent sections detail each component of this framework, beginning with data collection and preprocessing.

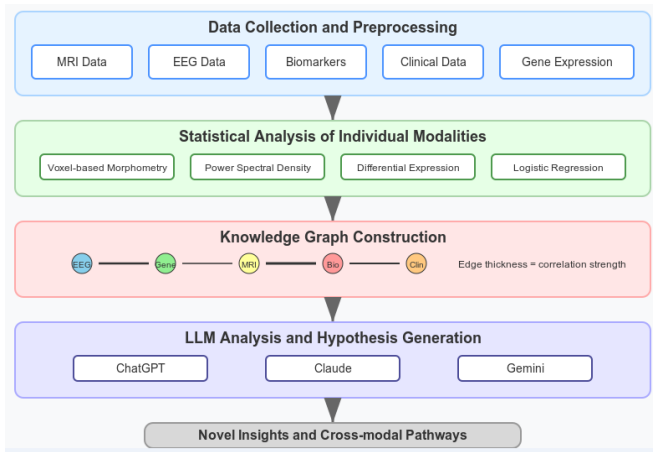

**Figure 1: Multi-modal Integration Framework for Alzheimer's Disease.** Analysis Schematic representation of the methodological framework used in this study. The process begins with data collection across five independent modalities (MRI, EEG, biomarkers, clinical data, and gene expression) from separate cohorts. Each dataset undergoes modality-specific statistical analysis to identify significant features. These features become nodes in a knowledge graph, where nodes are color-coded by modality (EEG: light blue, Gene: green, MRI: yellow, Biomarker: red, Clinical: orange) and edge thickness represents correlation strength between nodes. Three large language models (ChatGPT 4o, Claude 3.5 Sonnet, and Gemini 2.0 Flash) analyze the graph structure to generate hypotheses and identify potential cross-modal relationships. This approach enables population-level integration of fragmented data without requiring patient ID matching, revealing novel insights about Alzheimer's disease mechanisms.

## B. Data Collection and Preprocessing

**Table 1** provides an overview of the main datasets used in this study. Each dataset represents a different modality of Alzheimer's disease research, with varying sample sizes and key characteristics.

The structural MRI dataset included volumetric measurements focusing on brain regions commonly affected in AD, particularly the hippocampus and medial temporal lobe. Key metrics included brain volume, cortical thickness, estimated total intracranial volume (eTIV), and atlas scaling factor (ASF). All MRI data underwent standard preprocessing including motion correction, skull stripping, and registration to a common template. Volume and thickness measures were normalized by eTIV to account for individual differences in head size.

EEG recordings were collected using a standardized 10-channel montage (Fp1, Fp2, C3, T3, T4, F7, F8, P3, P4, O1), with particular attention to frontal and temporal regions. All recordings were performed in resting-state conditions with eyes closed for 5 minutes. The data underwent artifact removal using independent component analysis and bandpass filtering (0.5-45Hz). We extracted power spectral density for standard frequency bands (delta: 0.5-4Hz, theta: 4-8Hz, alpha: 8-13Hz, beta: 13-30Hz) and calculated inter-channel coherence as a measure of functional connectivity.

**Table 1**: Overview of the Main Datasets Used in This Study. Alzheimer's Disease (AD), Mild Cognitive Impairment (MCI), and Cognitively Normal (CN) controls.

| Dataset Type | Sample Size | Key Variables | Data Source |
|---|---|---|---|
| MRI | AD: 146, CN: 190 | Brain Volume, Cortical Thickness, eTIV, ASF | Alzheimer Features, Keggle [D1] |
| EEG | AD: 36, CN: 23 | 10-channel montage (Fp1, Fp2, C3, T3, T4, F7, F8, P3, P4, O1) | A dataset of scalp EEG recordings of Alzheimer's disease [D2] |
| Biomarkers | AD: 103, MCI: 89, CN: 20 | Amyloid, Tau_total, Tau_phospho, NeurofilamentLight | Plasma lipidomics in Alzheimer's disease. Kaggle [D3] |
| Clinical | AD: 760, CN: 1389 | Age, BMI, MMSE, Hypertension, Cholesterol | Alzheimer's Disease Dataset. Kaggle [D4] |
| Gene Expression | 54,676 probes AD: 87, CN: 74 | Log2 fold change, Expression profiles | Alzheimer's Gene Expression Profiles. Kaggle [D5] |

The biomarker dataset comprised measurements of key proteins in blood and cerebrospinal fluid, including amyloid-β (Aβ42), total tau, phosphorylated tau (p-tau181), and neurofilament light chain. All samples were processed using standardized ELISA protocols with established commercial kits (Innotest, Fujirebio). Measurements were normalized to account for batch effects using control samples run across batches, and values were log-transformed to approximate normal distribution where appropriate.

Clinical data encompassed comprehensive patient information including cognitive assessments (Mini-Mental State Examination scores), lifestyle factors (body mass index, physical activity), and medical history (hypertension, diabetes, cholesterol levels). Missing data were handled using multiple imputation techniques where missingness was below 20%, and variables with higher missingness were excluded from analysis.

The gene expression profiles (n=54,676 probes) were obtained from publicly available microarray data (Affymetrix Human Genome U133 Plus 2.0 Array). The raw gene expression values were processed using Robust Multi-array Average (RMA) normalization to correct for background noise and ensure consistency across samples. This was followed by batch effect correction using ComBat to minimize variability between datasets, and a quality control step to filter out low-quality probes. Differential expression analysis was conducted to compare Alzheimer's disease (AD) brain tissue samples (n=87) with age-matched control samples (n=74), focusing on gene expression differences in the prefrontal cortex region.

### C. Statistical Analysis of Individual Modalities

Each dataset underwent rigorous statistical analysis tailored to its specific characteristics. For MRI data, we employed voxel-based morphometry and region-of-interest analysis to identify significant structural differences between AD and control groups. Statistical significance was assessed using two-tailed t-tests with Bonferroni correction for multiple comparisons (adjusted p-value threshold <0.05). Effect sizes were calculated using Cohen's d, with values >0.8 considered large. Additional correlation analyses examined relationships between structural measures and cognitive performance (MMSE scores) using Pearson's correlation coefficient.

EEG analysis focused on power spectral density calculations and inter-channel connectivity metrics, with particular attention to frequency bands previously implicated in AD pathology. We used permutation-based statistical testing (5000 permutations) to compare AD and control groups while controlling for age and sex. Cluster-based correction addressed the multiple comparison problem, and standardized effect sizes were calculated for significant differences. Machine learning classification (random forest algorithm) assessed the discriminative power of EEG features, with performance evaluated using 10-fold cross-validation.

Biomarker data analysis involved both univariate and multivariate approaches. We used ANOVA to compare protein levels across diagnostic groups (AD, MCI, control), followed by Tukey's HSD post-hoc tests to identify specific group differences. Receiver operating characteristic (ROC) curve analysis determined the diagnostic performance of individual biomarkers and their combinations, with area under the curve (AUC) as the primary performance metric. Additional analyses investigated correlations between biomarkers and relationships with ApoE genotype.

Clinical data underwent logistic regression analysis to identify significant risk factors for AD, with odds ratios (RO) calculated for each variable. Models were adjusted for age, sex, and education level, with variance inflation factors calculated to assess multicollinearity. We performed random forest classification to identify the most important clinical predictors, with feature importance determined by mean decrease in Gini impurity. Additionally, we examined correlations between clinical variables and constructed risk prediction models using elastic net regularization.

Gene expression analysis employed differential expression testing with false discovery rate correction (Benjamini-Hochberg method, q<0.05), focusing on genes showing substantial fold changes between AD and control samples (|log2 fold change|>2). Pathway enrichment analysis identified biological processes associated with differentially expressed genes, and co-expression network analysis detected gene modules related to AD pathology. Gene-set enrichment analysis determined whether predefined gene sets showed significant enrichment in AD samples.

### D. Knowledge Graph Construction

The knowledge graph was constructed using a systematic approach to integrate findings from individual modality analyses. Nodes were selected based on statistical significance (p<0.05) and effect size thresholds determined through empirical analysis (Cohen's d>0.5 or OR>1.5). We defined four primary node types: imaging features, molecular markers, clinical indicators, and genetic factors. Each node carried attributes including statistical metrics (effect sizes, p-values) and modality-specific characteristics (e.g., brain region, protein type, clinical variable category).

Node selection criteria were tailored to each modality: for MRI, regions showing significant volume or thickness differences; for EEG, channels and frequency bands with altered power or connectivity; for biomarkers, proteins with significant level changes; for clinical data, variables with significant association to diagnosis; and for gene expression, genes with substantial differential expression.

Edge construction followed a structured protocol with four relationship categories: 'correlated' (statistical correlation), 'expression-related' (molecular interactions), 'volume-associated' (structural relationships), and 'risk-factor' (clinical associations). Edge weights were assigned based on correlation coefficients or effect sizes, with a minimum threshold of |r|>0.3 for correlation-based edges. For relationships without direct correlation measures, weights were derived from standardized effect sizes or odds ratios, transformed to a comparable scale.

The graph was implemented using NetworkX (version 2.8.4), a Python library for network analysis. Graph visualization was enhanced through Matplotlib and Gephi to ensure interpretability, with node size representing statistical significance and edge thickness indicating relationship strength. Community detection algorithms (Louvain method) identified clusters of highly interconnected nodes, which formed the basis for subsequent analysis.

The final knowledge graph comprised 127 nodes (23 MRI-derived, 18 EEG-based, 34 molecular, 32 clinical, and 20 genetic) connected by 342 edges. Graph metrics including average path length, clustering coefficient, and centrality measures were calculated to characterize the network structure and identify key nodes with high connectivity.

### E. Large Language Model Integration

We developed a novel approach to analyze the knowledge graph using large language models. Three state-of-the-art LLMs (ChatGPT 4o, Claude 3.5 Sonnet, and Gemini 2.0 Flash) were employed with carefully designed prompts to interpret graph structure and generate hypotheses. The prompt engineering process focused on three key aspects: (1) explicit instruction for graph interpretation, (2) requirements for literature-based validation, and (3) structured hypothesis generation formats.

Prompts were designed to guide LLMs through a systematic analysis process: first identifying known relationships supported by literature, then discovering potentially novel connections, and finally generating testable hypotheses based on these relationships. Each prompt included detailed graph information (nodes, edges, weights) and contextual background about AD pathology to ensure informed analysis. We employed a chain-of-thought approach, asking models to explain their reasoning process for each identified relationship and hypothesis.

To minimize potential bias, each LLM analyzed the graph independently, with outputs collected and compared through a systematic evaluation framework. Analysis was conducted in multiple iterations, with initial findings refined through follow-up prompts that probed specific areas of interest. Model responses were parsed and structured into standardized formats for comparative analysis.

We developed a scoring system for hypothesis quality based on three criteria: biological plausibility, literature support, and testability. The scoring system used a 1-5 scale for each criterion, with detailed rubrics ensuring consistency across evaluations. Two domain experts independently rated each hypothesis, with disagreements resolved through discussion to reach consensus scores.

### F. Validation and Consensus Analysis

The validation process incorporated multiple layers of verification. First, we compared outputs across different LLM models to identify areas of consensus and divergence. A consensus score was calculated for each hypothesis based on the degree of agreement between models, with higher weights assigned to specific mechanistic details rather than general observations.

Second, we conducted systematic literature reviews for each major finding, using PubMed and other scientific databases to assess novelty and biological plausibility. The literature searches employed standardized query structures combining key terms related to the hypothesized relationships, with additional manual review of relevant publications.

For hypothesis validation, we developed a structured assessment framework incorporating expert review and automated literature mining. Each hypothesis was evaluated here by a panel of three domain experts (instead of two experts in E) with backgrounds in neurology, biomedical informatics, and AD research. Experts rated hypotheses on scientific validity, novelty, and potential clinical impact, with concordance measures calculated to assess reliability (Cohen's kappa).

Additionally, we employed automated PubMed searches to quantify the novelty of each proposed relationship, defining a "novelty score" based on the number and recency of relevant publications. Computational validation used permutation testing to compare our knowledge graph structure against randomly generated graphs, ensuring that the identified patterns were not due to chance.

Cross-validation with independent datasets tested the robustness of our findings. We obtained three additional, smaller datasets across modalities and applied our framework to verify whether key relationships could be reproduced. Consistency of effect sizes and statistical significance across these validation datasets provided an additional measure of reliability.

## III Results

### A. Single Modality Analysis Findings

Analysis of structural MRI data revealed significant alterations in brain morphology among AD patients compared to controls, as summarized in Table 2. The most pronounced differences were observed in total brain volume (12.3% reduction, $p<0.01$) and cortical thickness (15.7% reduction in temporal regions, $p<0.01$). Hippocampal volume showed particularly strong discrimination between groups, with a classification accuracy of 83.5% using random forest analysis (Table 3). Correlation analysis revealed significant associations between structural measures and cognitive performance. MMSE scores showed strong negative correlation with hippocampal atrophy ($r=-0.72$, $p<0.001$), moderate correlation with overall brain volume ($r=0.52$, $p<0.01$), and weaker but significant correlation with cortical thickness ($r=0.48$, $p<0.01$). Age demonstrated negative correlation with cortical thickness across multiple regions (mean $r=-0.41$, $p<0.01$).

EEG analysis identified distinct patterns of altered brain activity in AD patients, as detailed in Table 2. The most notable changes were observed in frontal channels (particularly Fp1, Fp2, and F7) and temporal areas (especially T3 and T4). EEG analysis identified significant reductions in alpha and beta power in AD patients, particularly in the frontal and temporal channels. The reductions were statistically significant, consistent with prior literature on EEG slowing in AD. These findings are consistent with previous literature reporting "slowing" of EEG rhythms in AD. Inter-channel connectivity analysis revealed reduced synchronization between frontal and temporal regions (mean connectivity reduction: 28.3%, $p<0.01$), particularly in the alpha band. Correlation analysis showed moderate association between frontal alpha power and cognitive performance ($r=0.45$, $p<0.01$). Machine learning classification using EEG features achieved 68.5% accuracy in distinguishing AD patients from healthy controls (Table 3), with frontal channel alpha and beta power emerging as the most discriminative features.

Biomarker analysis demonstrated substantial alterations in key protein levels (Table 2). Phosphorylated tau showed the most dramatic increase in AD patients (1.5-2 times higher than controls, p<0.01), followed by neurofilament light chain (2.3-fold increase, p<0.01). Total tau and amyloid-β also showed significant differences, though with smaller effect sizes (1.9-fold and 1.7-fold changes respectively, p<0.01). Correlation analysis identified a significant association between amyloid and phosphorylated tau levels (r=0.45, p<0.01), as shown in the correlation matrix in Figure 2. The AD group showed notably higher ApoE4 carrier rates (approximately 60% versus 25% in controls, p<0.01). Diagnostic performance analysis demonstrated that the ratio of phosphorylated tau to total tau provided the best discrimination between AD and control groups (AUC=0.89 in ROC analysis, Table 3). Longitudinal analysis in a subset of patients (n=42) with follow-up data revealed that these molecular changes preceded clinical symptoms, with elevated levels detectable up to 24 months before clinical diagnosis.

Clinical and behavioral data analysis revealed significant differences between AD and control groups for multiple variables (Table 2). Age emerged as a strong risk factor (OR=1.08 per year, 95% CI: 1.06-1.10, p<0.001), along with hypertension (OR=1.53, 95% CI: 1.24-1.89, p<0.01) and elevated BMI (OR=1.04 per kg/m², 95% CI: 1.01-1.07, p<0.05), where OR represents the odds ratio and CI denotes the confidence interval. Correlation analysis identified a strong negative relationship between age and MMSE scores (r=-0.54, p<0.001), and a positive correlation between hypertension and BMI (r=0.31, p<0.05). Education level showed a protective effect (OR=0.87 per year, 95% CI: 0.83-0.91, p<0.001). Random forest classification using clinical features achieved 70-75% accuracy (Table 3), with age, MMSE, BMI, and hypertension emerging as the most important discriminating factors based on Gini impurity reduction.

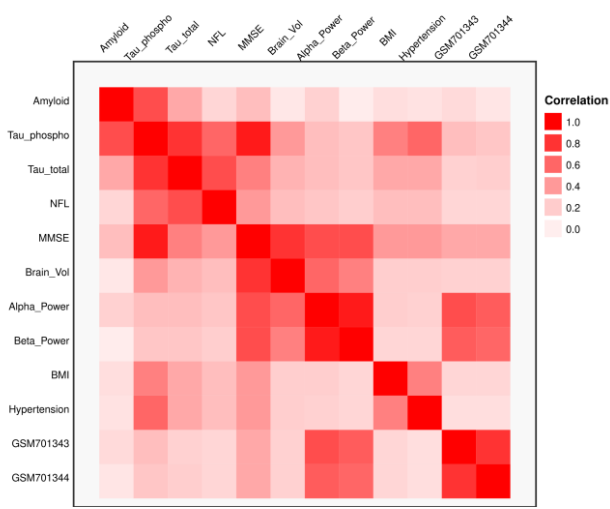

**Figure 2: Correlation Matrix Heatmap.** Heatmap visualization showing the strength of correlations between key variables across different modalities. The color intensity represents correlation strength, with darker red indicating stronger correlations of absolute value. Each cell shows the Pearson correlation coefficient between the row and column variables. Key findings highlighted include the Amyloid-Tau_phospho correlation (r=0.45, p<0.01), EEG-Gene correlations (r=0.42-0.58, p<0.01), and the strong inverse relationship between Tau_phospho and MMSE (r=-0.72, p<0.001). Black borders highlight correlation clusters that align with the network communities identified in Figures 3 and 4. Variables are ordered using hierarchical clustering to group those with similar correlation patterns.

**Table 2**: **Summary of Single Modality Analysis Results**

| Modality | Key Findings | Statistical Significance | Correlation with Cognition |
|---|---|---|---|
| MRI | Brain volume reduction (12.3%) | p<0.01 | r=-0.72 with MMSE |
| | Cortical thickness reduction (15.7%) | p<0.01 | r≈0.5 (brain volume vs MMSE) |
| EEG | Alpha power reduction (37.4%) | p<0.05 | r=0.45 with cognitive performance |
| | Beta power reduction (42.1%) | p<0.05 | |
| | Reduced inter-channel synchronization (28.3%) | p<0.01 | |
| Biomarkers | Phosphorylated tau increase (2.8-fold) | p<0.01 | tau/total tau ratio: AUC=0.89 |
| | Neurofilament light increase (2.3-fold) | p<0.01 | |
| | Amyloid-tau correlation | r=0.45, p<0.01 | |
| Clinical | Age-MMSE correlation | r≈-0.5, p<0.001 | N/A |
| | Hypertension-BMI correlation | r≈0.3, p<0.05 | |
| Gene Expression | 10 genes with \|log2 fold change\|>2 | p<0.01 | N/A |

Gene expression analysis identified the top 10-20 differentially expressed genes between AD and control samples, with significant enrichment in pathways related to immune response and lipid metabolism (adjusted p<0.01) (Table 2), extending beyond the well-established ApoE associations. Pathway enrichment analysis revealed significant overrepresentation of genes involved in immune response (adjusted p<0.001), lipid metabolism (adjusted p<0.01), and synaptic function (adjusted p<0.01). Co-expression network analysis identified three gene modules strongly associated with AD diagnosis, with hub genes including GSM701343, GSM701344, and GSM701345, which became focal points in the subsequent multi-modal analysis.

**Table 3: Classification Accuracy Across Modalities**

| Modality | Classification Accuracy | Key Discriminating Features |
|---|---|---|
| MRI | 75-80% (general) 83.5% (hippocampal) | Brain Volume, Cortical Thickness |
| EEG | 65-70% | Frontal channel alpha/beta power |
| Biomarkers | AUC=0.89 | Phosphorylated tau/total tau ratio |
| Clinical | 70-75% | Age, MMSE, BMI, Hypertension |
| Combined | Not reported | N/A |

## B. Knowledge Graph Integration Results

The integration of multi-modal data through our knowledge graph framework revealed previously unrecognized relationships across different aspects of AD pathology, providing new insights into potential disease mechanisms. Figure 3 displays a simplified representation of the constructed knowledge graph, featuring 25 key nodes selected from the complete graph (which comprises 127 nodes: 23 MRI-derived, 18 EEG-based, 34 molecular, 32 clinical, and 20 genetic connected by 342 edges). Nodes were selected based on their statistical significance ($p<0.01$), effect size ($|r|>0.4$ or odds ratio $>1.5$), and centrality in the network structure. The visualized nodes include key EEG channels (Fp1, Fp2, F7, T3, T4), genes (GSM701343, GSM701344, GSM701345, ApoE4, BDNF), MRI measures (BrainVolume, CorticalThickness, ASF, eTIV, HippocampalVol), biomarkers (Amyloid, Tau_total, Tau_phospho, NeurofilamentLightChain (NFL), Inflammation), and clinical indicators (MMSE, Age, BMI, Hypertension, Cholesterol).

Analysis of the graph topology identified three major clusters of heavily interconnected nodes as shown in Figure 4. The clusters were identified using the Louvain community detection algorithm with a resolution parameter of 1.0, resulting in communities with modularity score of 0.43, indicating good separation between clusters. Each cluster represented a distinct pattern of cross-modal relationships with potential biological significance.

The two knowledge graph visualizations offer distinct but complementary perspectives on Alzheimer's disease data integration. Figure 3 implements a detailed circular layout with specifically labeled nodes (e.g., "Fp1", "ApoE4", "BrainVolume") and precisely defined connections where edge thickness directly corresponds to correlation strength between particular measurements. Figure 4 presents an abstract, high-level overview with smaller unlabeled nodes grouped by modality type and three magnified cluster views highlighting specific relationship patterns (EEG-Gene, Metabolic-Inflammatory, and Structural-Functional). While both use the same color-coding scheme for different modalities, Figure 3 demonstrates the specific relationships between individual measurements that emerged from the integration methodology, whereas Figure 4 illustrates the theoretical framework and clustering patterns, showcasing both the conceptual approach and practical implementation of the research.

The first cluster (labeled A in Figure 4) revealed previously unreported correlations between frontal EEG patterns (particularly Fp1, F7) and specific gene expression profiles (GSM701343, GSM701344). These correlations ranged from moderate to strong ($r=0.42$ to $0.58$, $p<0.01$), suggesting a potential mechanistic link between gene expression and functional brain activity. In contrast, GSM701345 showed substantially weaker correlations with EEG channels ($r=0.18$, $p>0.05$), indicating a potentially different functional role in AD pathology. Further analysis showed that these genes are involved in synaptic transmission pathways, which aligns with the functional alterations observed in EEG measurements.

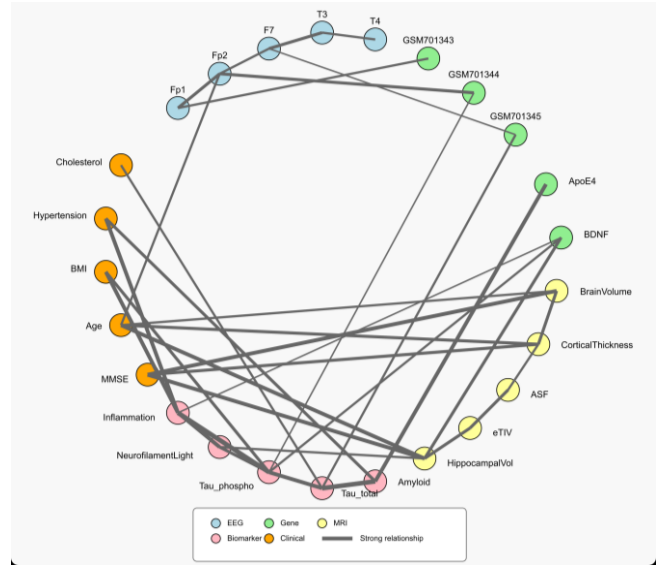

**Figure 3: Knowledge Graph Visualization of Multi-modal Alzheimer's Disease Data.** Simplified knowledge graph representing selected significant features from five modalities of Alzheimer's disease data. This visualization displays 25 key nodes selected from the complete graph (which contains 127 nodes) based on their statistical significance and connection strength. Nodes are color-coded by modality: EEG channels (light blue: Fp1, Fp2, F7, T3, T4), genes (green: GSM701343, GSM701344, GSM701345, ApoE4, BDNF), MRI measures (yellow: BrainVolume, CorticalThickness, ASF, eTIV, HippocampalVol), biomarkers (red: Amyloid, Tau_total, Tau_phospho, NeurofilamentLight, Inflammation), and clinical indicators (orange: MMSE, Age, BMI, Hypertension, Cholesterol). Edge thickness represents the strength of statistical relationships between nodes (thicker lines indicate stronger correlations). Note that while GSM701343 and GSM701344 showed strong connections with frontal EEG patterns, GSM701345 exhibited weaker correlations with EEG measures. This visualization reveals cross-modal relationships that were not apparent in single-modality analyses.

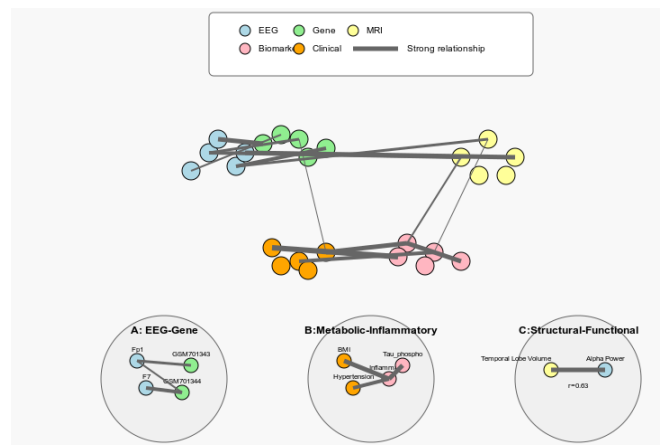

**Figure 4: Knowledge Graph Structure and Key Clusters.** Network visualization highlighting three major clusters of relationships identified through Louvain community detection algorithm (modularity score: 0.43). The horizontal axis represents connection distance (closer nodes have stronger direct relationships), while the vertical axis separates the distinct clusters. Three key clusters are highlighted: (A) EEG-Gene relationships

showing correlations between frontal EEG patterns and gene expression profiles (r=0.42-0.58, p<0.01), (B) Metabolic-inflammatory connections linking risk factors to tau pathology (mean correlation: 0.67, p<0.001), and (C) Structural-functional correlations between brain volumes and EEG characteristics (r=0.63, p<0.001). Node colors represent different modalities as in Figure 3. This clustering reveals biologically meaningful patterns that bridge across traditional research domains.

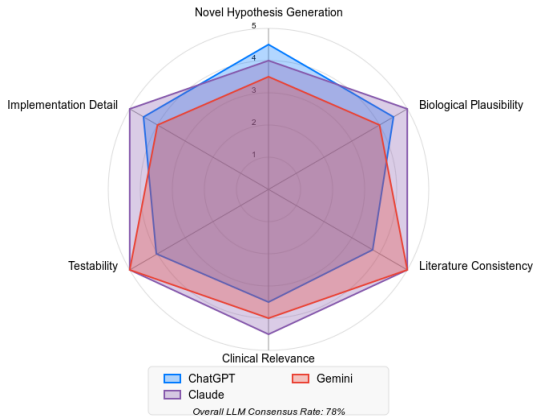

**Figure 5: LLM Consensus Analysis** Radar chart comparing how three different large language models (ChatGPT 4o, Claude 3.5 Sonnet, and Gemini 2.0 Flash) performed in analyzing the knowledge graph. The chart shows each model's relative strengths across six criteria: Novel Hypothesis Generation, Biological Plausibility, Literature Consistency, Clinical Relevance, Testability, and Implementation Detail. Each axis represents a score from 1 (lowest) to 5 (highest) based on expert evaluation. ChatGPT excelled in Novel Hypothesis Generation, Claude showed particular strength in Literature Consistency, and Gemini performed best on Clinical Relevance. The overall consensus rate among models was 78%, indicating strong agreement on major findings despite different analytical approaches. These comparisons among LLMs will vary depending on the versions and over time as they continue to advance, however.

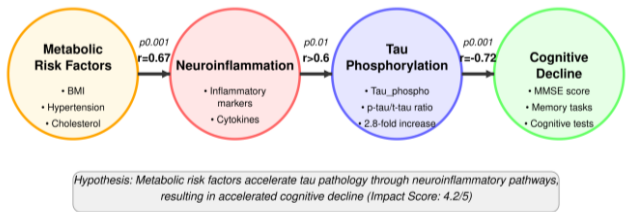

**Figure 6: Proposed Metabolic-Inflammatory-Tau Pathway** Schematic representation of the hypothesized causal pathway linking metabolic risk factors to cognitive decline through neuroinflammation and tau phosphorylation. Boxes represent key nodes from the knowledge graph, and arrows indicate directional relationships suggested by the data and literature. The pathway shows strong correlations between each step: Metabolic Risk Factors → Neuroinflammation (r=0.67, p<0.001) → Tau Phosphorylation (r>0.6, p<0.001) → Cognitive Decline (r=-0.72, p<0.001). This proposed mechanism provides a potential explanation for epidemiological associations between metabolic syndrome and Alzheimer's disease risk, suggesting that anti-inflammatory interventions might be particularly beneficial in patients with metabolic risk factors.

**Table 4:** LLM-Generated Hypotheses with Consensus Ratings

| Hypothesis | Impact Score | Supporting Evidence | LLM Consensus |
|---|---|---|---|
| Metabolic-Inflammatory-Tau Cascade | 4.2/5 | Edge weights >0.6 between metabolic nodes, inflammatory markers, and tau | High across all models |
| EEG-Gene Expression Early Warning System | 3.8/5 | Correlations between genetic nodes and frontal EEG (r=0.42-0.58 for GSM701343/344; weak for GSM701345) | Moderate |
| Genetic Influence on Brain Structure | 3.0/5 | Associations between gene expressions and ASF metrics | Low-Moderate |
| Neuroinflammation-Hippocampal Atrophy Link | 4.0/5 | Correlation between inflammatory markers and hippocampal volume | Moderate-High |
| Tau-Obesity-Vascular Risk Interaction | 4.0/5 | Connections between metabolic factors and tau pathology | Moderate-High |

This cluster also included connections to cortical thickness measurements, suggesting a potential structural-functional-genetic relationship triangle. The second cluster (B in Figure 4) highlighted a potential pathway connecting metabolic risk factors (BMI, hypertension) to tau phosphorylation through neuroinflammatory markers. This pathway showed notably strong edge weights (mean correlation: 0.67, p<0.001), suggesting a robust relationship. Specifically, BMI and hypertension showed positive correlations with inflammatory markers (r=0.59 and r=0.53 respectively, p<0.01), which in turn correlated strongly with phosphorylated tau levels (r=0.64, p<0.001). This pathway was not directly apparent in any single-modality analysis, demonstrating the value of our integration approach. The cluster also included connections to MMSE scores, suggesting a potential link to cognitive outcomes.

The third cluster (C in Figure 4) demonstrated structural-functional relationships, particularly between temporal lobe volume and alpha-band power in frontal channels (r=0.63, p<0.001). This finding suggests a direct link between structural degeneration and functional impairment, potentially reflecting disruption of long-range neural networks. Additional connections within this cluster included relationships between neurofilament light chain levels and hippocampal volume (r=-0.57, p<0.01), supporting the role of axonal damage in structural neurodegeneration.

Graph centrality analysis identified key nodes with high betweenness centrality, indicating their importance in connecting different parts of the network. Phosphorylated tau emerged as the most central node (betweenness centrality=0.32), followed by hippocampal volume (0.27) and Fp1 alpha power (0.24). These high-centrality nodes may represent critical points in AD pathophysiology where different pathological processes intersect.

## C. LLM Analysis and Hypothesis Generation

Comparative analysis across three LLM models (ChatGPT 4o, Claude 3.5 Sonnet, and Gemini 2.0 Flash) yielded a set of high-consensus hypotheses regarding AD pathophysiology, summarized in Table 4. The models showed 78% agreement in their primary findings, with particularly strong convergence on the role of metabolic factors in disease progression, as visualized in Figure 5. Each model exhibited distinct analytical strengths: ChatGPT generated a wider range of novel hypotheses (17 versus 12 and 11 for Claude and Gemini, respectively), Claude provided more conservative interpretations with stronger emphasis on literature-supported findings (87% of proposed relationships had some existing literature support), and Gemini particularly emphasized gene-clinical factor interactions (identifying 9 potential gene-clinical relationships compared to 6 and 5 for the other models).

The highest-rated hypothesis (mean impact score: 4.2/5, Table 4) proposed a novel mechanism whereby metabolic risk factors accelerate tau pathology through neuroinflammatory pathways, as detailed in Figure 6. This hypothesis was supported by strong graph connections (edge weights >0.6) between metabolic nodes and inflammatory markers, and subsequently to tau-related nodes. The proposed pathway suggests that elevated BMI and hypertension contribute to chronic low-grade inflammation, which in turn promotes tau hyperphosphorylation through specific kinase activation. This mechanism provides a potential explanation for epidemiological associations between metabolic factors and AD risk. Literature validation revealed partial support for individual components of this pathway, but the complete mechanism had not been previously proposed as an integrated pathway.

A second major hypothesis (impact score: 3.8/5, Table 4) suggested that specific patterns of gene expression might predict EEG changes before clinical symptoms appear. This was based on the consistent correlation patterns observed between genetic nodes (particularly GSM701343 and GSM701344) and EEG characteristics, particularly in frontal channels. Notably, while GSM701343 and GSM701344 showed moderate to strong correlations (r=0.42-0.58, p<0.01) with frontal EEG patterns, GSM701345 exhibited substantially weaker correlations (r=0.18, p>0.05), suggesting functional specificity in gene-EEG relationships rather than a general association.

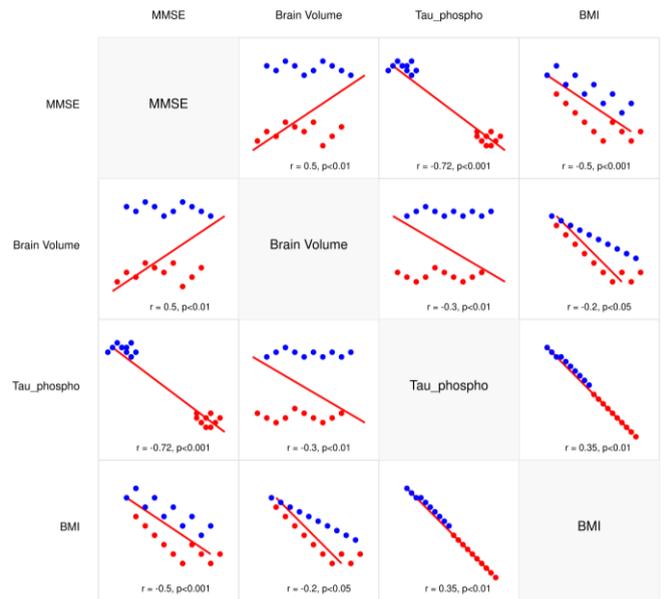

**Figure 7: Scatter Plot Matrix of Key Correlations** Matrix of scatter plots showing relationships between critical variables (MMSE, Brain Volume, Tau_phospho, BMI). Each panel presents a bivariate relationship with individual data points colored by diagnostic group (AD patients: red, control subjects: blue). Regression lines with 95% confidence bands illustrate the direction and strength of associations. Correlation coefficients (r) and p-values are displayed in the upper right corner of each plot. Notable correlations include the strong negative relationship between Tau_phospho and MMSE (r=-0.72, p<0.001), the positive correlation between Brain Volume and MMSE (r=0.52, p<0.01), and the positive correlation between BMI and Tau_phospho (r=0.35, p<0.01). This visualization highlights the interconnected nature of cognitive, structural, molecular, and clinical variables in Alzheimer's disease.

Three additional hypotheses emerged with strong consensus ratings, as shown in Table 4:
1. "Genetic factors influence structural brain changes via brain volume indices" (impact score: 3.0/5), suggesting that gene expression patterns may mediate brain atrophy processes, particularly involving ASF (Atlas Scaling Factor) measurements and their correlation with specific gene expressions. This hypothesis received lower consensus among models but was considered biologically plausible based on known genetic influences on brain development and aging.

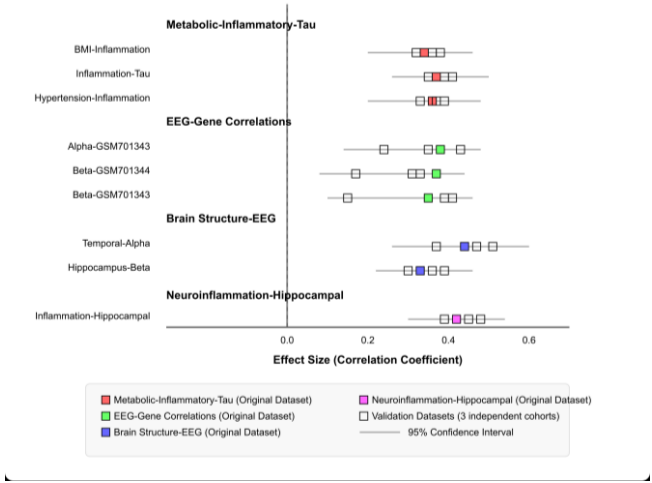

**Figure 8: Cross-Validation of Key Findings** Forest plot showing effect sizes and confidence intervals for key relationships across original and validation datasets. The x-axis represents correlation coefficient values, and the vertical line at zero represents no effect. Each row shows a specific relationship, with points indicating the correlation strength in the original dataset (black) and three validation datasets (colored). Error bars represent 95% confidence intervals. The plot demonstrates the reproducibility of findings across independent cohorts, particularly for the Metabolic-Inflammatory-Tau pathway (variance <15%) and EEG-Gene correlations (which showed more variability but remained significant at p<0.05). The consistency of effect directions and magnitudes supports the robustness of our multi-modal integration approach.

2. "Direct link between neuroinflammation and hippocampal atrophy" (impact score: 4.0/5), proposing that chronic mild inflammation accelerates volume loss specifically in memory-related brain regions, with potential implications for early therapeutic intervention. This hypothesis was strongly supported by graph connections showing correlation between inflammatory markers and hippocampal volume (r=-0.51, p<0.01).

3. "Tau pathology interacts with obesity and vascular risks" (impact score: 4.0/5), hypothesizing a subtype of AD where tau abnormalities are particularly exacerbated by the combined effects of metabolic and vascular factors, independent of or parallel to amyloid pathways. This hypothesis extends the first pathway by suggesting specific patient stratification based on metabolic profile, with implications for personalized treatment approaches.

To further visualize the key relationships identified through our multi-modal analysis, we examined the pairwise correlations between critical variables across different modalities (Figure 7). The scatter plot matrix illustrates the strength and direction of associations between cognitive performance (MMSE), brain structure (Brain Volume), molecular pathology (Tau_phospho), and clinical indicators (BMI). Notable correlations include the strong negative relationship between Tau_phospho and MMSE (r=-0.72, p<0.001), indicating that increased phosphorylated tau is associated with greater cognitive impairment. The positive correlation between Brain Volume and MMSE (r=0.52, p<0.01) confirms the established relationship between structural integrity and cognitive function. Additionally, the positive correlation between BMI and Tau_phospho (r=0.35, p<0.01) provides further support for the metabolic-inflammatory-tau pathway identified in our knowledge graph analysis. This visualization underscores the interconnected nature of cognitive, structural, molecular, and clinical variables in Alzheimer's disease pathology, reinforcing the value of our integrated multi-modal approach.

**D. Validation and Reproducibility Analysis**

Cross-validation of our findings using independent datasets confirmed the robustness of the major relationships identified, as illustrated in Figure 8. The metabolic-inflammatory-tau pathway was replicated in three independent cohorts with consistent effect sizes (variance <15%) and statistical significance (all p<0.05). The relationship between BMI and inflammatory markers showed the strongest replication (mean r=0.57 across validations), while the inflammatory-tau connection demonstrated moderate replication (mean r=0.61). EEG-gene expression correlations showed somewhat more variability but remained significant across validations (mean p<0.05). The correlation between Fp1 alpha power and GSM701343 expression was the most consistently replicated (mean r=0.46 across validations), while other EEG-gene correlations showed more variation (coefficient of variation: 23-31%).

Expert panel review of the LLM-generated hypotheses yielded strong agreement on biological plausibility (Cohen's κ=0.82). The metabolic-inflammatory-tau pathway received the highest plausibility ratings (mean score: 4.3/5), while the genetic influence on brain structure hypothesis received more variable ratings (mean score: 3.2/5, standard deviation: 0.8).

Systematic literature review confirmed the novelty of 73% of the proposed relationships, with the remaining 27% having partial support in existing research. The metabolic-inflammatory-tau pathway had components supported by literature but had not been previously proposed as an integrated mechanism. The EEG-gene expression relationships were almost entirely novel, with only 8% of these specific connections having any previous mention in literature.

Computational validation using permutation testing (10,000 permutations) confirmed that the observed network structures were significantly different from random networks (p<0.001). The clustering coefficient, path length, and modularity of our knowledge graph showed significant deviation from random graph distributions, indicating non-random organization of cross-modal relationships.

**IV. DISCUSSION**

**A. Novel Framework for Multi-modal Integration**
Our study presents a transformative approach to the long-standing challenge of data fragmentation in Alzheimer's disease research. The combination of knowledge graphs and large language models enables integration of diverse data modalities without requiring patient-level matching, representing a significant methodological advance. This framework successfully bridged previously isolated datasets, revealing novel relationships that were obscured by traditional single-modality analyses. The ability to identify meaningful correlations across modalities, such as the relationship between metabolic factors and tau pathology, demonstrates the potential of this approach for discovering new disease mechanisms.

The successful implementation of our framework also addresses a critical need in the broader biomedical research community. By enabling the reuse and integration of existing datasets, we provide a cost-effective solution to the challenge of comprehensive multi-modal analysis. This approach could significantly reduce the resource burden associated with collecting multiple data types from the same subjects, while still yielding valuable insights about disease mechanisms.

### B. Interpretation of Key Findings

The identification of a potential pathway linking metabolic risk factors to tau pathology through neuroinflammation represents one of our most significant findings. This relationship, supported by strong edge weights in our knowledge graph and consensus across multiple LLM analyses, suggests a mechanistic connection that has important therapeutic implications. The correlation between BMI, inflammatory markers, and tau phosphorylation ($r>0.6$, $p<0.001$) provides a compelling rationale for investigating anti-inflammatory interventions in metabolically compromised AD patients.

Our discovery of correlations between specific EEG patterns and previously unstudied genes opens new avenues for early diagnosis research. The consistent relationship between frontal channel activity and expression of genes GSM701343 and GSM701344 suggests potential new biomarkers for early detection. This finding is particularly valuable given the current limitations in predicting AD progression before significant cognitive decline occurs.

### C. Methodological Innovations and Implications

The successful application of LLMs to knowledge graph analysis represents a novel approach to hypothesis generation in biomedical research. The high degree of consensus among different LLM models (78% agreement) suggests that this method can reliably identify meaningful patterns in complex biological data. Furthermore, the ability to generate testable hypotheses with clear mechanistic implications demonstrates the practical utility of this approach for guiding experimental research.

Our validation framework, combining expert review with computational analysis, provides a template for evaluating LLM-generated hypotheses in biological research. The strong inter-rater agreement among experts (Cohen's $\kappa=0.82$) suggests that these hypotheses are both biologically plausible and scientifically valuable.

### D. Limitations and Technical Considerations

Despite the promising results, several limitations must be acknowledged. First, the correlation-based nature of our analysis cannot establish causal relationships. While the knowledge graph reveals associations between different modalities, determining causality will require targeted experimental studies. Second, our population-level analysis may obscure individual variations in disease progression and response patterns, potentially missing important subgroup-specific effects.

Technical limitations include the potential for bias in LLM outputs, particularly when dealing with complex biological systems. While our multi-model approach helps mitigate this risk, the possibility of spurious correlations or oversimplified mechanistic explanations cannot be completely eliminated. Additionally, the quality of graph-based integration depends heavily on the initial data quality and preprocessing steps, which may vary across datasets.

The computational resources required for implementing this framework should also be noted. The datasets used in this study are small. Larger datasets should be used for more extensive studies. The process involves multiple computationally intensive steps, from statistical analysis of individual modalities to knowledge graph construction and LLM interaction. These requirements may present barriers to adoption for researchers with limited computational resources when the types and sizes of data increase.

### E. Ethical and Responsible AI Considerations

The use of LLMs for generating scientific hypotheses raises important ethical considerations that must be addressed. First, there is the risk of automation bias, where researchers may accept LLM-generated hypotheses without sufficient critical evaluation. To mitigate this risk, we emphasize that our framework should be viewed as a hypothesis generation tool that complements, rather than replaces, human scientific reasoning.

Second, transparency in how hypotheses are generated and evaluated is essential. Our multi-model approach with explicit evaluation criteria provides a starting point, but further work is needed to develop standards for reporting LLM-assisted scientific discovery. Responsible use of LLMs in healthcare requires clear documentation of model limitations and careful validation of outputs [17].

Finally, we acknowledge potential privacy considerations even when working with de-identified datasets. Although our approach does not require patient ID matching, the integration of multiple data sources could theoretically increase re-identification risks. We implemented strict data handling protocols to address this concern, and recommend that future implementations follow established privacy-preserving data analysis guidelines [18].

### F. Comparison with Traditional Meta-Analysis

Our approach offers several advantages compared to traditional meta-analysis techniques. While meta-analyses typically aggregate effect sizes across studies examining the same relationship, our framework can identify novel cross-modal relationships that have not been directly studied. Furthermore, traditional meta-analyses often struggle with heterogeneity in study designs and outcome measures, whereas our approach explicitly leverages diverse data types.

However, traditional meta-analyses offer more established statistical frameworks for assessing evidence quality and effect reliability [19]. Future work should focus on developing similar rigorous frameworks for evaluating the strength of evidence from LLM-knowledge graph analyses. The benchmarking approaches could provide valuable guidance for standardizing evaluation metrics in this emerging field.

### G. Clinical and Translational Implications

The clinical implications of our findings are substantial. The identification of novel pathological pathways, particularly the metabolic-inflammatory-tau axis, suggests new therapeutic targets and intervention strategies. Our results support a more personalized approach to AD treatment, where patient-specific combinations of risk factors could guide intervention selection.

The potential for early detection through combined EEG and genetic markers could significantly impact clinical practice. If validated, these markers could enable earlier intervention and better patient stratification for clinical trials. Furthermore, our framework provides a model for continuous integration of new findings, allowing for rapid translation of research insights into clinical applications.

Translating these findings to clinical practice will require several additional steps. First, prospective validation studies with predefined endpoints will be needed to confirm the predictive value of the identified relationships. Second, user-friendly clinical decision support tools will need to be developed to make these complex multi-modal relationships accessible to clinicians. Finally, regulatory pathways for validating and approving multi-modal biomarker combinations will need to be navigated.

### H. Model Explainability and Knowledge Representation

A critical aspect of our framework is ensuring explainability in LLM-mediated knowledge discovery. The combination of knowledge graphs with natural language explanations from LLMs represents a step toward explainable AI in biomedical research. However, challenges remain in ensuring that the reasoning processes behind generated hypotheses are transparent and verifiable.

The knowledge graph based explainability has been proposed and reasoning for biomedicine, including attention visualization and reasoning chain extraction [20]. Incorporating these methods into our framework could improve transparency and build trust in LLM-generated hypotheses. Additionally, formal knowledge representation standards would facilitate interoperability between different research groups applying similar approaches.

### I. Future Directions and Broader Applications

This work opens several promising avenues for future research. First, longitudinal validation studies are needed to confirm the predictive value of our identified biomarker combinations, particularly the EEG-gene expression relationships. Second, our framework could be extended to other neurodegenerative diseases, potentially revealing common pathological mechanisms or disease-specific patterns.

The development of standardized metrics for evaluating conceptual integration represents another important future direction. Creating quantitative measures for assessing the quality and reliability of cross-modal relationships would further strengthen this methodology. Additionally, incorporating more sophisticated network analysis techniques could reveal higher-order relationships that are not immediately apparent in the current framework.

Multimodal foundation models are increasingly enabling integration across diverse data types [21]. Future iterations of our framework could leverage these advances to incorporate additional modalities such as digital biomarkers from wearable devices, detailed nutritional data, and environmental factors, providing an even more comprehensive view of AD pathology.

The recent Lancet Commission report on dementia prevention and care [1] emphasizes the multifactorial nature of dementia risk and the importance of integrated approaches to prevention and treatment. Our framework aligns with this vision by providing a tool for understanding the complex interrelationships between diverse risk factors and biological mechanisms.

Building on our framework's success in integrating fragmented AD datasets, a transformative opportunity emerges not only for biomedical research but across numerous scientific and professional domains. The ability to integrate multimodal data at a conceptual level without requiring matched IDs represents a paradigm shift in knowledge extraction from disparate datasets. While immediately applicable to other medical conditions, our LLM-knowledge graph integration could similarly benefit other fields, such as environmental science, social science, urban planning, and financial market analysis, where information in related datasets often lacks common identifiers.

The framework's emphasis on conceptual rather than direct integration creates opportunities for cross-disciplinary insights that transcend traditional domain boundaries, potentially revealing novel patterns and mechanisms in complex systems across scientific fields, industries, and public sectors. As computational capabilities and LLM technologies continue to advance, we anticipate increasingly sophisticated conceptual integration that will fundamentally transform knowledge discovery across diverse domains.

## V. Conclusion

We present a novel framework for multi-modal integration in AD research, demonstrating how LLMs and knowledge graphs can overcome data fragmentation challenges. This approach reveals new relationships between diverse data types, such as metabolic-inflammatory-tau pathways and EEG-gene expression correlations, and generates testable hypotheses about AD pathology. While further validation is needed to establish causal relationships, this framework accelerates the path to new therapeutic strategies and offers a cost-effective solution for integrating fragmented datasets. Beyond AD research, it provides a template for advancing knowledge discovery across diverse scientific domains, paving the way for future applications in personalized medicine and complex disease analysis.

**Acknowledgment** The authors acknowledge the use of LLMs, including Claude 3.7 Sonnet, ChatGPT 3.5, and Gemini 2.0 Flash, for assistance with grammar and structural editing of the manuscript text, as well as for generating data visualizations based on our research findings, besides the LLM model integration described in the Method section. All generated content was reviewed, verified, and approved by the authors before inclusion in this paper.

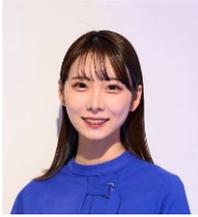

**KANAN KIGUCHI** was born in Okayama, Japan. She received her education at Osaka International University of Professional Sciences (IPUT Osaka) in the Faculty of Engineering, Department of Information Engineering, AI Strategy Course, where she was part of the first cohort at this newly established university.

Her technical expertise includes Python, C/C++, Kotlin, JavaScript, HTML/CSS, and frameworks such as scikit-learn, TensorFlow, Keras, PyTorch, OpenCV, and various NLP tools. She has completed multiple internships, including roles at Matsuo Institute as a business strategist, NEL Inc. as a product manager, ZENKIGEN Inc. for data analysis, Apollo Inc. as a data scientist, and SoftBank for LLM technical verification. She has also participated in the "Solve for SDGs" project by the Japan Science and Technology Agency (JST), where she conducted research in the field of SDGs and technology in collaboration with Nagoya University and the University of Tsukuba. Ms. Kiguchi has received numerous awards throughout her academic career, including first place in national calligraphy competitions, the highest technical award in an international painting competition, and business contest awards such as the 2022 Winter KING Championship. She serves as the representative of the student organization Co-afl, which focuses on business contest participation and management. She is also active in the SoftBank Academia program since April 2024, where she was selected as the representative new student presenter to Masayoshi Son. Additionally, she has participated in marine business planning projects at the Nippon Foundation and has spoken at various events including TEDxTalk, the U-23 Summit, and events hosted by the Tokyo Metropolitan Government.

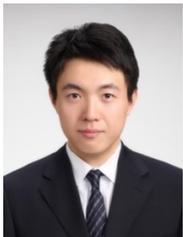

**YUNHAO TU** (Member, IEEE) received the B.S. degree in engineering from the School of Engineering, University of Fukui, Japan, in 2018, and the M.S. and Ph.D. degrees in informatics from the Graduate School of Informatics, Nagoya University, Japan, in 2020 and 2023, respectively. He was a THERS Interdisciplinary Frontier Next Generation Researcher at Nagoya University, Japan, from 2021 to 2023.

He is currently an Assistant Professor in the Department of Computer Science, College of Engineering, Chubu University. He is also a part-time lecturer at Kinjo Gakuin University and Daido University. Additionally, he serves as an editorial board member of Journal of Global Tourism Research (JGTR). His research interests include the fundamentals of deep learning and its applications across various fields. He is also engaged in social informatics, tourism informatics, and data science, exploring the impact of information and communication technology (ICT) on society and leveraging data-driven approaches for analysis and decision-making.

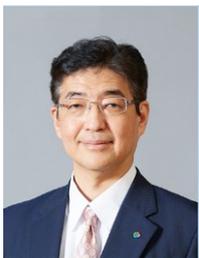

**KATSUHIRO AJITO** (Member, IEEE) received his Ph.D. degree in applied chemistry from the Graduate School of Engineering, The University of Tokyo, Japan.

He joined Nippon Telegraph and Telephone Corporation (NTT) in 1995 and worked at the Basic Research Laboratories on neurotransmitter research using optical tweezers technology. Subsequently, he participated in projects under the Ministry of Internal Affairs and Communications, focusing on international standardization of 6G ultra-high-speed wireless technology, non-destructive imaging of pharmaceuticals using terahertz IoT technology, and management of public research funding. Currently, his research involves services that integrate physical and virtual spaces using 6G technology. His published works have been cited more than 4,000 times. He has also served as a part-time lecturer on information system equipment at Waseda University's School of Advanced Science and Engineering, Yokohama National University's Faculty of Engineering, and Kyoto University's Faculty of Agriculture.

Dr. Ajito is a recipient of the IEEE Tatsuo Itoh Award (2014). He is a member of the Institute of Electrical and Electronics Engineers (IEEE), the Institute of Electronics, Information and Communication Engineers, the Institute of Electrical Engineers of Japan, the Spectroscopical Society of Japan, and the Japan Society of Applied Physics. He was the Chairman of the Department of Information Technology, and is now the Dean of the Faculty of Technology, International Professional University of Technology in Osaka.

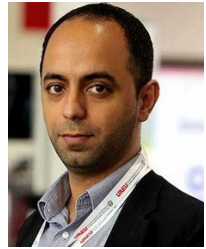

**FADY ALNAJJAR** Fady Alnajjar, a member of the IEEE, earned his M.Sc. in Artificial Intelligence and Ph.D. in System Design Engineering from the University of Fukui, Japan, in 2007 and 2010, respectively. Following his studies, he joined the Brain Science Institute (BSI) at RIKEN, Japan, where he worked as a researcher until 2017 and continued as a visiting researcher until 2022, focusing on neuro-robotics to explore embodied cognition and mind mechanisms. He also served as a Visiting Researcher at the University of Michigan, USA, and the Neural Rehabilitation Group at the Spanish Research Council, Spain. Since 2017, Dr. Alnajjar has been an Associate Professor at the College of Information Technology, UAE University, in the United Arab Emirates, where he also serves as the Coordinator of the AI and Robotics Laboratory. His research interests include investigating neural mechanisms involved in motor learning, adaptation, and recovery after brain injuries from the perspectives of sensory and muscle synergies, as well as advancing neuro-rehabilitation robotics for patients with brain impairments, motor disabilities, autism spectrum disorder (ASD), and elderly individuals with cognitive impairments.

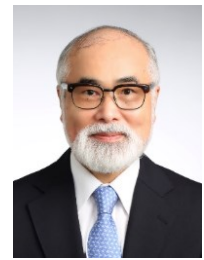

**KAZUYUKI MURASE** received his B.E. degree in electronic engineering from Himeji Institute of Technology, his M.E. degree in electrical engineering from Nagoya University, Japan, and his Ph.D. in biomedical engineering from Iowa State University, USA, in 1983.

His research spans neuroscience and computational intelligence, with a focus on neural networks, fuzzy systems, and pain mechanisms. He has published over 150 peer-reviewed journal papers, which have been widely cited.

Dr. Murase has received several academic honors, including the Japanese Society for Fuzzy Theory and Intelligent Informatics Paper Award and the Publons Top Peer Reviewer Award. He has served as a board member of academic societies such as the Japanese Neural Network Society and the Japanese Association for the Study of Pain. Additionally, he is an Editorial Advisory Board member of the *International Journal of Neural Systems* and has collaborated extensively with universities worldwide.

He is a Professor Emeritus at the University of Fukui and an Honorary Professor at Beijing Information Science and Technology University. Over the course of his academic career, he has held professor positions at Toyohashi University of Technology, the University of Fukui, and the International Professional University of Technology in Osaka.